\documentclass[
]{ceurart}

\begin{document}

\copyrightyear{2021}
\copyrightclause{Copyright for this paper by its authors.
  Use permitted under Creative Commons License Attribution 4.0
  International (CC BY 4.0).}

\conference{Forum for Information Retrieval Evaluation, December 15-18, 2023, India}

\title{Overview of the CLAIMSCAN-2023: Uncovering Truth in Social Media through Claim Detection and Identification of Claim Spans}

\author[1]{Megha Sundriyal}[%
email=meghas@iiitd.ac.in
]
\address[1]{IIIT Delhi, India}

\author[1]{Md Shad Akhtar}[%
email=shad.akhtar@iiitd.ac.in,
]

\author[2]{Tanmoy Chakraborty}[%
email=chak.tanmoy.iit@gmail.com,
]
\address[2]{IIT Delhi, India}

\begin{abstract}
A significant increase in content creation and information exchange has been made possible by the quick development of online social media platforms, which has been very advantageous. However, these platforms have also become a haven for those who disseminate false information, propaganda, and fake news. Claims are essential in forming our perceptions of the world, but sadly, they are frequently used to trick people by those who spread false information. To address this problem, social media giants employ content moderators to filter out fake news from the actual world. However, the sheer volume of information makes it difficult to identify fake news effectively. Therefore, it has become crucial to automatically identify social media posts that make such claims, check their veracity, and differentiate between credible and false claims. In response, we presented CLAIMSCAN in the 2023 Forum for Information Retrieval Evaluation (FIRE'2023). The primary objectives centered on two crucial tasks: Task A, determining whether a social media post constitutes a claim, and Task B, precisely identifying the words or phrases within the post that form the claim. Task A received 40 registrations, demonstrating a strong interest and engagement in this timely challenge. Meanwhile, Task B attracted participation from 28 teams, highlighting its significance in the digital era of misinformation.

\end{abstract}


\maketitle

\section{Introduction}
The rapid growth of online social media platforms has facilitated a significant increase in content creation and information exchange, which has been highly beneficial. However, these platforms have also become a breeding ground for those who spread malicious rumors, fake news, propaganda, and misinformation. Claims play a vital role in shaping our understanding of the world, but unfortunately, they are often used by purveyors of fake news to deceive people. The COVID-19 \textit{``Infodemic"} is a prime example of this phenomenon, which has resulted in the widespread dissemination of false information about politics and social issues, as well as fake medical claims \cite{naeem2020covid}. To address this issue, social media giants hire content moderators to separate fake news from the real thing. However, the sheer volume of information makes it difficult to identify fake news effectively. As a result, automatically identifying posts on social media platforms containing such claims, verifying their validity, and distinguishing between credible and false claims has emerged as a critical research problem in NLP.

\begin{table}[t]
    \centering
    \caption{Task A: Representative examples of claims and non-claims.}
    \scalebox{1}
    {
    \begin{tabular}{p{36em}c}
    \toprule
         \multicolumn{1}{c}{\bf Text} & \bf Claim\\         
         \midrule  
         
         My heartfelt gratitude goes out to the men and women in uniform who did not back down from putting their lives in danger to save the lives of our citizens in difficult circumstances. & No \\ \midrule
         
         According to research into the dangers of cooking with aluminum foil, some of the toxic metal can contaminate food. This is especially true when cooking or heating spicy or acidic foods in foil. Aluminum levels in the body have been linked to osteoporosis and Alzheimer's disease. & Yes \\ \midrule
         
         Furthermore, health insurers should recognize alternative medicine as a treatment option because there is a chance of recovery. & No \\ \midrule
         
         Toothpaste Zaps Pimples. Don't pop your pimples! Daily Glow recommends applying toothpaste to a pimple before bed and washing it off with warm water when you wake up in the morning. Toothpaste draws impurities out of pores while also drying the skin and shrinking the pimple. & Yes \\ 
         \bottomrule
    \end{tabular}
    }
    \label{tab:taska_examples}
    \end{table}

\begin{table}[t]
    \centering
    \caption{Task B: Representative examples of claims and their claim spans.}
    \scalebox{1}
    {
    \begin{tabular}{p{27em}p{12em}}
    \toprule
         \multicolumn{1}{c}{\bf Claim} & \multicolumn{1}{c}{\bf Claim Span}\\         
         \midrule  
         % RT @username: No no no. Corona beer is the cure not the disease. & Yes \\ \midrule
                  
         According to research into the dangers of cooking with aluminum foil, some of the toxic metal can contaminate food. This is especially true when cooking or heating spicy or acidic foods in foil. Aluminum levels in the body have been linked to osteoporosis and Alzheimer's disease. & cooking with aluminum foil, some of the toxic metal can contaminate food. \\ \midrule
                  
        Toothpaste Zaps Pimples. Don't pop your pimples! Daily Glow recommends applying toothpaste to a pimple before bed and washing it off with warm water when you wake up in the morning. Toothpaste draws impurities out of pores while also drying the skin and shrinking the pimple. & Toothpaste Zaps Pimples.  \\
        \bottomrule
    \end{tabular}
    }
    \label{tab:taskb_examples}
    \end{table}

The concept of a claim, defined by \citet{toulmin2003uses} as an assertion that deserves attention, is central to Argument Mining (AM). However, the segregation of claims is complex and challenging due to language structure and context variation across different sources. Differentiating between claims and non-claims is highly subjective and tricky, making it difficult for human annotators and advanced state-of-the-art neural models. Table \ref{tab:taska_examples} furnishes a few examples of claims and non-claims for more understanding. Although claim-detecting systems have advanced, there is still room for improvement in their precision and efficiency \cite{sundriyal2021desyr}. The dynamic nature of online social media platforms presents a significant challenge. New types of misinformation can emerge quickly, and keeping up with changing trends and patterns can take time. In addition to the challenges of efficiently identifying claims, another factor affecting the fact-checking task is extracting precise snippets of the claim from the entire social media post, which often contain extraneous irrelevant text \cite{sundriyal-etal-2022-empowering}. Table \ref{tab:taskb_examples} depicts claims and their corresponding claim spans. 

Disentangling such argumentative units of misinformation from benign statements has numerous advantages, including performing downstream tasks like claim check-worthiness and verification, adding explainability to the coarse-grained claim detection task, and simplifying the fact-checking process for human fact-checkers. This task, however, is complex and requires overcoming technical obstacles such as language complexity and variability. 

To this end, we present the CLAIMSCAN-2023, a shared task in the 2023 edition of the Forum for Information Retrieval Evaluation workshop. Through this shared task, we aim to develop systems that can effectively detect and identify claims within social media text. To accomplish this, we propose two sub-tasks:  
\begin{itemize}
    \item Task A Claim Detection: Given a social media post, the task is to identify whether or not the post contains a claim. 
    \item Task B Claim Span Identification: Given a social media post containing a claim, the objective is to pinpoint the exact phrase of the post that constitutes the claim. 
\end{itemize}

\section{Background}
\label{sec:related-work}
The growth of online social media has greatly amplified the spread of misinformation, primarily through disseminating false claims. This presents a significant risk to online users, as misinformation can spread rapidly without any effective countermeasures in place. Consequently, tasks related to identifying and handling claims have gained considerable prominence within the field of Natural Language Processing (NLP), particularly as a crucial precursor to automated fact verification. Claims, as a core component of misinformation, have been the subject of extensive research from multiple perspectives in recent years. This includes areas such as Claim Detection  \cite{chakrabarty2019imho, gupta2021lesa, sundriyal2021desyr}, Claim Check-worthiness \cite{jaradat2018claimrank, wright2020claim}, Claim Span Identification \cite{sundriyal-etal-2022-empowering, mittal2023mcsi}, Claim Normalization \cite{sundriyal2023claimnorm}, and Claim Verification \cite{zhi2017claimverif, hanselowski2018ukp, soleimani2020bert, sundriyal2022document}.

Pioneering efforts in the study of claims can be attributed to \citet{bender-etal-2011-annotating}, who introduced the ``Authority and Alignment in Wikipedia Discussions" corpus, which comprised around 365 discussions sourced from Wikipedia Talk Pages. This work garnered substantial attention from researchers focusing on claims and served as the cornerstone for the challenging field of automated claim detection. Over the last decade, the investigation of online claims has gained some traction within the NLP research community. A primary attempt was made by \citet{rosenthal2012detecting}; they used a supervised approach based on sentiment and word-gram derivatives to mine claims from discussion platforms. Despite the fact that their work was limited to traditional machine-learning approaches, it laid the groundwork for future research in this field. Following research on claim detection, linguistically motivated features such as sentiment analysis, syntax, context-free grammar, and parse trees were heavily emphasized \cite{levy2014context, lippi2015context, levy2017unsupervised}. 

Given that the majority of studies at the time focused on domain-specific formal texts, \citet{daxenberger-etal-2017-essence} addressed this limitation by conducting cross-domain claim detection across six diverse datasets, revealing both distinctive and shared features across different domains. Recent research has led to the use of Large Language Models (\textit{LLMs}), which hold great promise. \citet{chakrabarty2019imho} demonstrated the power of fine-tuning with their ULMFiT language model, which was fine-tuned on a large Reddit corpus of approximately 5 million opinionated claims. A generalized claim detection model was proposed by 
\citet{gupta2021lesa} that detects the presence of a claim in any online text, regardless of source. They worked with both structured and unstructured data by training a combination of linguistic encoders (part-of-speech and dependency trees) and a contextual encoder. Because language models incur significant computational overheads, \citet{sundriyal2021desyr} addressed this issue and proposed a lighter framework that attempted to generate discernible feature spaces for individual classes while avoiding using LLMs and focusing on the definition-centric approach. Several computational social science researchers have expressed interest in the  \textit{CLEF-$2020$} shared task organized by the \textit{CheckThat! Lab} \cite{barron2020checkthat}. \citet{williams2020accenture} won the task by fine-tuning the RoBERTa model \cite{liu2019roberta}, which was further strengthened by mean pooling and dropout. With their RoBERTa vectors supplemented with Twitter meta-data, \citet{nikolov2020team} bagged second position.

The existing body of claim detection research primarily focuses on identifying claims at the sentence level rather than delving into the finer details of exact claim spans. As a result, a recent advancement in this field has moved away from broad, sentence-level claim identification models and toward more detailed, fine-grained claim span identification \cite{sundriyal-etal-2022-empowering}. The idea of rationales was first presented by \citet{zaidan2007using}, who highlighted segments of the text that validated the conclusions of their label.  They reported a significant improvement in performance after incorporating these rationales into the training process for sentiment classification of movie reviews. In the field of argumentation mining, \citet{trautmann2020fine} released the \textit{AURC-$8$} dataset, which includes token-level span annotations for the argumentative components of stance, as well as their corresponding label. The \textit{SemEval} community has initiated coarse-grained span identification concerning other domains of argument mining such as toxic comments \cite{pavlopoulos2021semeval} and propaganda techniques \cite{da-san-martino-etal-2020-semeval}. These shared tasks amassed many solutions constituting transformers \cite{chhablani2021nlrg}, convolutional neural networks \cite{coope2020span}, data augmentation techniques \cite{rusert2021nlp_uiowa, palliser2021hle, plucinski2021ghost}, and ensemble frameworks \cite{zhu2021hitsz, nguyen2021s}. \citet{wuhrl2021claim} compiled a corpus of around 1200 biomedical-related tweets with claim phrases. Apart from English, argument extraction has also been examined for other languages like Greek \cite{goudas2014argument, sardianos2015argument} and German \cite{habernal2017argumentation}. In a recent study conducted by \citet{sundriyal-etal-2022-empowering}, a systematic approach was presented for identifying claim spans within social media posts. Additionally, they created an extensive Twitter corpus manually annotated specifically for this task.

\section{Tasks Description and Settings}
\label{sec:tasks-description}

CLAIMSCAN-2023 shared task consists of two sub-tasks: Claim Detection and Claim Span Identification. Participants were free to engage in one or both sub-tasks. 

\paragraph{\textbf{Task A (Claim Detection):}} Given a social media post, the objective is to identify whether a claim is present within a provided post or not. This task can be quite demanding, as claims exhibit diverse structures and can be concealed within extensive text segments. Hence, the system needs to discern patterns and linguistic cues that are indicative of claims, which may encompass assertive language, explicit statements on a topic, and allusions to supporting evidence or sources. 

\paragraph{\textbf{Task B (Claim Span Identification):}} Following the initial determination of whether the post contains a claim, the subsequent step entails pinpointing the precise span of the claim within the post. It is crucial for the system to precisely identify the specific words or phrases that form the claim, as this information plays a pivotal role in assessing its accuracy during the fact-checking process.

\section{Datasets}
\label{sec:datasets}
To accomplish Task A (Claim Detection), we utilize a publicly available large-scale claim detection dataset developed and curated for tweets \cite{gupta2021lesa}. The dataset was manually annotated extensively using carefully crafted guidelines, yielding a collection of $9,894$ tweets labeled as either containing a claim or not containing a claim. The statistics of the dataset are detailed in Table \ref{tab:lesa-stats}. For Task B (Claim Span Identification), we use the CURT dataset, which contains $9,458$ claim spans from $7,555$ tweets \cite{sundriyal-etal-2022-empowering}. Table \ref{tab:curt-stats} contains the dataset statistics and details. This dataset has also been annotated manually, with each span identified and tagged using the \textit{BIO} (Begin-Inside-Outside) encoding scheme \cite{ramshaw-marcus-1995-text}, as shown in Table \ref{tab:curt-examples}. This tagging scheme indicates whether each word in the tweet is within a claim span and, if so, whether at the start or end of the span.

\begin{table}[h]
\centering
\caption{Task A: Statistics of claim detection dataset.} 
\scalebox{0.90}{
\begin{tabular}{lcc}
\toprule
\bf Dataset & \bf Claim & \bf Non-claim   \\ 
\midrule
\bf Train set & 7354 &	1055 \\ 
\bf Test set & 1296 &  189  \\ 
\midrule
\bf Overall & 8650 & 1244 \\
\bottomrule
\end{tabular}}
\label{tab:lesa-stats}
\end{table}

\begin{table}[h]
    \centering
    \caption{Task B: Statistics of claim span identification dataset. } 
    \scalebox{1}{
    \begin{tabular}{lccc}
    \toprule
      \bf Dataset & \bf Train &\bf Test &\bf Validation\\ 
        \midrule
        \bf  Total no. of claims &6044 &755 &756 \\
        \bf Avg. length of tweets &27.40 &26.93 &27.29 \\
        \bf Avg. length of spans &10.90 &10.97 &10.71 \\
        \bf No. of span per tweet &1.25 &1.20 &1.27 \\
        \bf No. of single span tweets &4817 &629 &593 \\
        \bf No. of multiple span tweets &1201 &121 &161 \\
        \bottomrule
        
    \end{tabular}
    }
    \label{tab:curt-stats}
\end{table}

\begin{table}[h]
    \centering
    \caption{A few examples of social media posts from CURT dataset \cite{sundriyal-etal-2022-empowering} and their corresponding \textit{BIO} tags depicting claim spans.}
    \label{tab:curt-examples}
    \scalebox{1}
    {
    \begin{tabular}{p{25em}p{14em}}
    \toprule
         \textbf{Text} & \textbf{Span} \\         
         \midrule
        
        @mcford77 @floradoragirl Exactly. that is the point. \textcolor{blue}{Home Schooling prevents loads of \#Coronavirus deaths.} & \textit{\{O, O, O, O, O, O, O, B, I, I, I, I, I, I\}}  \\ 
        \midrule 
        
        @JoeySalads Zero. \textcolor{blue}{\#Covid19 is a hoax}. \textcolor{blue}{The dead people died of something else.} Where are the rest of the Corpses? If \#coronavirus is real, then NYC would not be the greatest hit spot of DEATH from it in the world by a factor of five. What about Mexico City? Sydney? & \textit{\{O, O, B, I, I, I, B, I, I, I, I, I, I, O, O, O, O, O, O, O, O, O, O, O, O, O, O, O, O, O, O, O, O, O, O, O, O, O, O, O, O, O, O, O, O, O, O, O, O, O\}} \\ 
        \bottomrule
    \end{tabular}}
\end{table}

We took great care in developing annotation guidelines for both tasks, which went through several iterations and have already been published in two highly regarded peer-reviewed conferences. In addition, to ensure the quality of the data, we conducted pilot studies and enlisted human annotators with a strong understanding of claims and who are active social media users to manually annotate the datasets. This rigorous process helps to ensure data accuracy and reliability, resulting in more robust and reliable models. More details about the datasets can be found in \citet{gupta2021lesa} and \citet{sundriyal-etal-2022-empowering}.

\section{Evaluation Metrics}
\label{sec:evaluation-metrics}
The evaluation metric for both tasks is the F1 score. For Task A, we compute {\bf Macro-F1} scores using \textit{Scikit-learn} Library in Python used by the existing systems for claim detection \cite{gupta2021lesa, sundriyal2021desyr, chakrabarty2019imho}. For Task B, as the final labels for spans follow the \textit{BIO} tagging notation, our task becomes a sequence labeling task. We compute \textbf{Token-F1} scores following existing span detection methods \cite{pavlopoulos2021semeval, sundriyal-etal-2022-empowering}. Each team was allowed a maximum of 10 submissions, and the best scores obtained on test data were used for the leaderboard.

\section{Participating Systems and Results}
\label{sec:results}

Task A received 40 registrations, and Task B received 28 registrations. Out of these 6 teams submitted their official runs for Task A, while 4 submitted for Task B. We first describe the teams that submitted system description papers. 

\begin{itemize}
    \item \textbf{Team NLytics\cite{NLytics}:} Team NLytics participated in both subtasks. For Task A, they fine-tuned the RoBERTa model \cite{liu2019roberta} using RoBERTaForSequenceClassification, optimizing it with a regression loss (Binary Cross-Entropy Loss). They employed the AdamW optimizer with an initial learning rate of 2e-5. The optimizer followed a schedule where the learning rate increased linearly from 0 to the initial rate during a warm-up period and then decreased linearly to 0. The training process encompassed 20 training epochs. In Task B, they utilized RoBERTa and added a layer of linear-chain Conditional Random Field (CRF) \cite{lafferty2001conditional}. As RoBERTa operates with byte pair encoding (BPE) units, while CRF requires whole words, only the initial tokens of words were used as input to the CRF, with any word continuation tokens being excluded. The training was started with 20 epochs, with an early stopping callback monitoring the model's performance on the validation set.

    \item \textbf{Team mjs227\cite{mjs227}:} Team mjs277 participated only in Task B. For identifying claim spans, they used the positional transformer architecture. The positional transformer is a transformer encoder architecture variant that uses a position-sensitive attention mechanism called positional attention. The underlying language model in their proposed model was RoBERTa$_{BASE}$ \cite{liu2019roberta}. 

    \item \textbf{Team CODE\cite{NLytics}:} Team CODE participated in both subtasks. In Task A, they fine-tuned a BERT-based model \cite{devlin2018bert} optimized for sequence classification and trained for 5 epochs. They utilized a binary cross-entropy loss (BCE loss) and employed an Adam optimizer for this task. In Task B, they employed the RoBERTa model and conducted fine-tuning to predict a binary label (0 or 1) for every token, indicating whether the token is associated with a claim or not. Instead of using the IOB tag set, they adhered to IO tags. Their model underwent training for a duration of 4 epochs, and to eliminate noise, they excluded instances with claim spans consisting of fewer than three words.

\end{itemize}

\begin{table}[h]
{
\caption{Task A results for the best run per team based on macro-F1 scores.}
\centering
\begin{tabular}{ccc}
\toprule
\textbf{Rank} & \textbf{Name} & \textbf{Macro-F1} \\
\hline
1 & NLytics & 0.7002 \\
2 & bhoomeendra & 0.6900 \\
3 & amr8ta & 0.6678 \\
4 & CODE & 0.6526 \\
5 & michaelibrahim & 0.6324 \\
6 & pakapro & 0.4321 \\
\bottomrule
\label{tab:task_a_results}
\end{tabular}
}
\end{table}

The official results for Task A are presented in Table \ref{tab:task_a_results}. Among the six participating teams, Team NLytics clinched the top position, attaining a noteworthy macro-F1 score of $0.7002$. Following closely, Team bhoomeendra secured the second position.\footnote{They did not release their system description papers. \label{fn:no-submission}} Second position was bagged by Team amr8ta.\footref{fn:no-submission} In the fourth spot for Task A was Team CODE, achieving a macro-F1 score of $0.6526$. The fifth and sixth positions were occupied by Team michaelibrahim and Team pakapro, with macro-F1 scores of $0.6324$ and $0.4321$, respectively.\footref{fn:no-submission} It's worth noting the substantial margin between the top-performing team and the rest.

The official results for Task B are in Table \ref{tab:task_b_results}. Team mjs277 achieved the highest ranking among all participating teams, with a token-F1 of $0.8344$. To identify claim spans, they harnessed the positional transformer architecture, resulting in a substantial enhancement of their model's performance. Team bhoomeendra secured the second position in the task, achieving a token-F1 score of $0.8030$\footref{fn:no-submission}. Team NLytics attained the third spot by fine-tuning a RoBERTa model for predicting BIO tags for each token in the input sentence, complementing it with a Conditional Random Field (CRF) layer. In fourth place was Team CODE, who opted for \textit{IO} tags instead of \textit{BIO} tags to signify whether a token was part of the claim or not.

\begin{table}
{
\caption{Task B results for the best run per team based on token-F1 scores.}
\centering
\begin{tabular}{ccc}
\toprule
\textbf{Rank} & \textbf{Name} & \textbf{Token-F1} \\
\hline
1 & mjs227 & 0.8344 \\
2 & bhoomeendra & 0.8030 \\
3 & NLytics & 0.7821 \\
4 & CODE & 0.5714 \\
\bottomrule
\label{tab:task_b_results}
\end{tabular}
}
\end{table}

\section{Conclusion}
\label{sec:conclusion}
We presented the first edition of the CLAIMSCAN-2023 shared task. This shared task encompassed two vital subtasks within the fact-checking process, ranging from detecting claims in social media posts to determining the exact claim spans. These tasks collectively contribute to developing technology that aids human fact-checkers in their endeavors. We witnessed significant participation, with Task A drawing 40 registrations and Task B garnering 28 registrations. A total of 6 teams and 4 teams submitted official runs for Tasks A and B, respectively. We discussed the tasks and main findings of the three participating teams who submitted their systems based on their system description papers. We look forward to enriching our datasets with more examples, diverse information sources, and languages. Our overarching objective is to share our insights and inspire researchers to bridge the gaps in the field, ultimately enhancing the effectiveness of fact-checking systems and contributing to a safer online environment. In the future, we also aim to expand the scope of our task to encompass a broader range of modalities, such as images.

%% Define the bibliography file to be used
\bibliography{sample-ceur}

\appendix

\end{document}